\documentclass[runningheads]{llncs}
\usepackage[T1]{fontenc}

\usepackage{color}
\usepackage{tabularx,multirow,graphicx,float,subfig}
\usepackage{manyfoot}
\DeclareNewFootnote{A}[*]
\DeclareNewFootnote{B}[]

\begin{document}
\title{A Statistical Analysis of LLMs' Self-Evaluation Using Proverbs}

\author{Ryosuke Sonoda * \inst{1} and Ramya Srinivasan * \inst{2}}

\institute{Fujitsu Limited, \email{sonoda.ryosuke@fujitsu.com}  \and Fujitsu Research of America, 
\email{ramya@fujitsu.com}}

\maketitle              
\def\thefootnote{*}\footnotetext{equal contribution}
\def\thefootnote{}\footnotetext{A poster of this paper will be presented in PRICAI'24}
\def\thefootnote{\arabic{footnote}}
\begin{abstract}

Large language models (LLMs) such as ChatGPT, GPT-4, Claude-3, and Llama are being integrated across a variety of industries. Despite this rapid proliferation, experts are calling for caution in the interpretation and adoption of LLMs, owing to numerous associated ethical concerns. Research has also uncovered shortcomings in LLMs' reasoning and logical abilities, raising questions on the potential of LLMs as evaluation tools. In this paper, we investigate LLMs' self-evaluation capabilities on a novel proverb reasoning task. We introduce a novel proverb database consisting of 300 proverb pairs that are similar in intent but different in wordings, across topics spanning gender, wisdom, and society. We propose tests to evaluate textual consistencies as well as numerical consistencies across similar proverbs, and demonstrate the effectiveness of our method and dataset in identifying failures in LLMs' self-evaluation which in turn can highlight issues related to gender stereotypes and lack of cultural understanding in LLMs. 
\keywords{Large language models  \and Bias \and Statistical analysis.}
\end{abstract}

\section{Introduction}
There has been a lot of excitement surrounding large language models (LLMs) with some reckoning that these could been seen as early versions of artificial general intelligence (AGI) systems by virtue of their mastery over a wide array of novel and difficult tasks \cite{bubeck2023}. Recent studies also argue that LLMs like ChatGPT exhibit human-like intuitive behavior, avoiding some common reasoning biases that are found in other LLMs \cite{thilo2023}.

Despite these promising findings and applications, numerous studies have shed light on a variety of errors, biases, stereotypes, and other concerns associated with ChatGPT and its advanced versions \cite{brown,liang,pan,laura,ji_2022,zhou_2023}. Calling for caution in interpreting the capabilities of generative AI as akin to human intelligence, the authors in \cite{west2024} demonstrate that although models can outperform humans in generation, they consistently fall short of human capabilities in measures of understanding. Studies have shown that ChatGPT and GPT4 do not perform well on logical reasoning tasks, especially on out-of-distribution and natural language inference datasets \cite{liu2023,bian2023,adrian2023,arkoudas2023}. 

Aforementioned drawbacks become more pronounced in the context of cultural understanding where nuanced and subtle aspects of non-Western cultures are largely misunderstood by LLMs \cite{yan,tarek}. For example,  recent research has shown that cross-lingual consistency of factual knowledge transfer in various multilingual pre-trained language models is largely biased towards English \cite{jirui}. Studies have also emphasized on the need to have balanced multilingual pre-training datasets to better represent the diversity of different cultures with many implications on the topic of cross-lingual transfer \cite{badr}.

We complement and extend previous works on critical analysis of LLMs by introducing a novel proverb reasoning task in which LLMs are asked to evaluate their own content by providing a textual reasoning of two similar proverbs and corresponding numerical scores that would reflect the appropriateness of the reasoning. Some prior works have analyzed for semantic consistency in content generated by LLMs, uncovering hallucinations and other such issues of concern \cite{manakul,shu,peiyi,harsh}. Unlike these  works that have focused mostly on examining textual consistencies, in this work, we propose metrics for examining score consistencies in addition to textual consistencies of the generated responses. Towards achieving this objective, we introduce a new dataset consisting of proverb pairs for three topics---gender, wisdom, and society, each topic containing about a hundred proverbs pairs. 

To construct the proverb based prompts, we use publicly available sources such as \cite{kaggle,epic-proverbs,am,wikiquote,ucla} and manually construct a corresponding proverb for each original proverb such that it is similar in intent, but slightly different in wordings. For example, the proverb prompts concerning gender topic would only differ in the gender word such as in  {\it `what does a poor workman always blames his tools mean?'} vs {\it`what does a poor workwoman always blames her tools mean?'}. For the topic on wisdom, the proverbs based prompts are constructed such that the rationale remains unaltered. An example pair in this topic from our dataset--- {\it  `what does it mean to say a bird in hand is worth two in a bush? }, vs {\it `what does it mean to say a bird in hand is worth three in a bush?'} The proverb pairs in society topic mostly reflect similar proverbs across different geographies/cultures on various aspects of people and communities---for example--- {\it `what does the hyena calls another hyena worse than itself mean?'} vs {\it `what does the pot calling the kettle black mean?'} 

For every proverb in the dataset, we obtain five textual responses from LLMs explaining the rationale behind the proverb along with a numerical score for the same, thus resulting in ten responses and ten scores for each proverb pair. Since the proverbs in a proverb pair pertain to similar contexts (albeit slight changes that would not alter the overall intent of the proverb), one would expect to observe consistencies in scores and textual responses across the corresponding pairs. We utilize natural language inference based methods to compute average entailment scores for measuring textual consistency \cite{manakul,maynez,williams-etal-2018-broad}. We leverage the non-parametric Siegal-Tukey test \cite{erich} to examine for consistency in numerical scores as well as in the computed textual consistency scores. 

The contributions of the paper can be summarized as follows. 

1) We introduce a novel proverb reasoning task for analyzing consistency in LLMs' textual explanations and numerical evaluation. Towards this, we introduce a novel dataset consisting of 100 proverb pairs each across three topics: gender, wisdom, and society, resulting in a total of 600 proverbs. The proverb pairs are constructed such that they connote similar contexts but differ in literal wordings, thereby serving as a testbed for identifying inconsistencies in generated evaluations. \\
2) We propose the use of natural language inference based methods and non-parametric Siegal-Tukey statistical test for analyzing score consistencies and textual consistencies across similar contexts. We demonstrate how these tests can be leveraged for detecting failures in LLMs' reasoning and numerical evaluation that may in turn reflect gender based stereotypes, lack of common sense reasoning, as well as a lack of cultural understanding. \\

\section{Dataset}
\vspace{-0.3in}
\begin{table} [h]
\caption{Types of question prompt templates for topics.} 
\label{tab4}
\centering
\begin{tabular}{ll}
\hline
Topic & Question type  \\
\hline \hline
\multirow{1}{*}{Gender}
 & what does <PROVERB> mean? \\ 
 & why is it said that <PROVERB>? \\ \hline
\multirow{1}{*}{Wisdom}
 &  what does it mean to say <PROVERB> ? \\ 
 & give me a sentence appropriate to justify that <PROVERB>. \\ 
 & give me a proverb similar in meaning to <PROVERB>. \\ 
 & give me a proverb contrary in meaning to <PROVERB>. \\ 
 & what does <PROVERB> mean? \\ 
 & why is it said that <PROVERB>? \\ \hline
\multirow{1}{*}{Society}
 & what does <PROVERB> mean? \\ 
 & why is it said that <PROVERB>? \\ \hline
\end{tabular}
\end{table}

\begin{table} [h] 
\caption{Illustrations from the dataset for the topics on gender, wisdom, and society.  Q1: prompt question with an original proverb, Q2: prompt question with a manually constructed similar proverb.}\label{tab123}
\begin{tabularx}{\linewidth}{lX|X}
\hline 
Topic & Q1 &  Q2  \\
\hline \hline
\multirow{1}{*}{Gender}
& why is it said that adversity makes a man, wise not rich? &  why is it said that adversity makes a woman, wise not rich? \\ \cline{2-3}
& what does there is no strong  man  when the sea is at its worst mean? & what does there is no strong woman when the sea is at its worst mean? \\ \cline{2-3}
& why is it said that wise men think alike? & why is it said that wise women think alike? \\ \cline{2-3}
& what does every man has his price mean? & what does every woman has her  price mean? \\ \cline{2-3}
& why is it said that clothes maketh the man? & why is it said that clothes  maketh the woman? \\ \hline 
\multirow{1}{*}{Wisdom}
& what does the early bird catches the worm mean? &  what does the early crow catches the bug mean? \\ \cline{2-3}
& give me a proverb contrary in meaning to faith will move mountains & give me a proverb contrary in meaning to faith will move oceans\\  \cline{2-3}
& give me a proverb contrary in meaning to a miss is as good as a mile& give me a proverb contrary in meaning to a miss is as good as a kilometer\\  \cline{2-3}
& what does one flower does not bring spring mean?&  what does two flowers do not bring spring mean ?\\ \cline{2-3}
& why is one never too old to learn?& why is age no barrier to learning?\\ \hline 
\multirow{1}{*}{Society}
& what does the squirrel walks in the dew mean? &  what does the early bird catches the worm mean? \\ \cline{2-3}
& what does the mouth sells the head mean?  & what does the tongue talks at the head’s cost mean? \\  \cline{2-3}
& why is it said another’s mouth cannot take the oath for you? & why is it said every bird must hatch its own egg? \\  \cline{2-3}
& what does the pot cooks the food and does not eat it mean?&  what does bees that make honey do not taste it mean?\\ \cline{2-3}
& why is it said one knows a field of millet from its crop?& why is it said a tree is known by its fruits?\\\hline
\end{tabularx}
\end{table}

Given that LLMs are highly advanced and are known to perform well in Western dominated contexts and languages, it becomes important to probe them with texts that embed latent meanings, reflecting diverse cultures. It is well-known that proverbs embed latent meanings and are usually applicable across geographies and cultures \cite{peter}. For these reasons, we choose proverb databases to investigate failure modes of LLMs. We choose proverbs focused on three topics, namely, gender, wisdom, and society so as to be able to identify potential issues related to gender stereotyping, lack of wisdom, and lack of cultural understanding in LLMs. 

Each topic in our dataset consists of about 100 proverb pairs, resulting in a total of 600 proverbs across the three topics. As described in the introduction, to construct a proverb pair, we source a proverb from existing resources such as \cite{kaggle,epic-proverbs,am,wikiquote,ucla} and manually construct a corresponding proverb such that it is similar in intent to the original proverb but slightly different in literal wordings. For the topic of gender, we construct proverb pairs by changing all the gender words, as we would like to understand if the reasoning of LLMs are consistent across genders. For the topic on wisdom, we construct proverb pairs such that the rationale behind the proverb remains the same. For example, consider a proverb {\ A bird in hand is worth two in a bush}. Changing two to three should not alter the reasoning behind the proverb. So, we construct prompt pairs as follows---- {\it what does a bird in hand is worth two in a bush mean?} and {\it what does a bird in hand is worth three in a bush mean?}. For the topic on society, in order to examine if LLMs can understand cultural differences, we choose proverb pairs across different cultures that connote similar meanings \cite{am,ucla} and examine if evaluations are consistent. Although the prompts in our dataset are largely of the form {\it `what does <PROVERB> mean'?} or {\it `why is it said that <PROVERB>'?}, we do have a few proverb prompts of other types (see Table ~\ref{tab4}). Examples from our dataset can be found in Table~\ref{tab123}. 

\section{Method} 

We examine whether the numerical scores assigned by LLMs to their own responses are consistent across similar scenarios using two statistical tests.

\textbf{Notation.} Let $q$ represent a question based on an original proverb, and $q'$ represent a question based on a proverb with an equivalent meaning but expressed differently from the original proverb.

An LLM generates five answers, $R=\{r_1, r_2, r_3, r_4, r_5 \}$ and $R^{\prime}=\{r_1^{\prime}, r_2^{\prime}, r_3^{\prime}, r_4^{\prime}, r_5^{\prime} \}$, for $q$ and $q'$, respectively. Additionally, the LLM 
assigns scores, $S=\{s_1, s_2, s_3, s_4, s_5 \}$, and $S^{\prime}=\{s_1^{\prime}, s_2^{\prime}, s_3^{\prime}, s_4^{\prime}, s_5^{\prime} \}$, for $R$ and $R'$, respectively.

\subsection{Evaluating score consistency}
In order to evaluate score consistency, we want to analyze whether there are significant differences in the score pair 
$S$ and $S'$. 
Here, we assume that a pair of scores has a scoring error if there are significant differences in scores for question pairs where similar responses are expected.
Towards this, we employ the Siegal-Tukey (ST) test, which is a non-parametric statistical method to test the null hypothesis ($H_0$) that two independent scores come from the same population (e.g., a proverb premise) against the alternative hypothesis ($H_1$) that the samples come from populations differing in variability or spread. Thus,
$H_0$ : $\sigma_S$= $\sigma_S'$ and $MeS= MeS'$ and $H_1$ : $\sigma_S \geq \sigma_{{S}^{'}}$, where $\sigma$ and $Me$ are the variance and medians for 
S and S'. The test is entirely distribution-free and is particularly applicable in settings where the number of instances in a group is small ($5$ in our case).

\subsection{Evaluating text consistency}

Similar to the test for scores, we examine for differences in textual consistency between the two responses $R$ and $R'$.
Textual consistency has been considered in many existing studies~\cite{manakul,maynez,williams-etal-2018-broad}, with findings primarily relating consistency to textual response inaccuracy. 
We prefer to find a relationship between textual consistency and scoring failure. 
To do so, we assume that if there is a difference in textual consistency between unique proverb pairs, 
then that 
pair has a scoring failure.
This assumption is natural---if textual consistencies differ for questions that should ideally yield similar responses, it can indicate potential incorrect scoring by LLMs

To quantify textual consistency, we employ a method based on Natural Language Inference (NLI). NLI typically uses an LLM-based classifier that outputs the probability of entailment, neutral, and contradiction for the premise of hypothesis. In this paper, we use DeBERTa-v3-large~\cite{Pengcheng} fine-tuned to NLI~\footnote{\url{https://huggingface.co/potsawee/deberta-v3-large-mnli}} tasks which has shown promising results for estimating textual consistency~\cite{manakul}.
In our case, we consider whether a response $r_i$ is supported by other four responses and take the average entailment probability as the textual consistency $\mathrm{TC}_i$ of the response $r_i$. 
That is, 
\begin{equation}\label{eq:method2}
    \mathrm{TC}_i = \frac{1}{4} \sum_{j\neq i} P(\mathrm{entailment}\mid r_i, r_j), 
\end{equation}
where $P(\mathrm{entailment}\mid r_i, r_j)$ is the probability of entailment given the $r_i$ and $r_j$ as hypothesis and premise.  
Similarly, we calculate text consistency for $R'$ and obtain five $\mathrm{TC}'_i$.

Finally, we perform Siegal-Tukey test on five $\mathrm{TC}_i$ and five $\mathrm{TC}'_i$.

\section{Experiments}
\vspace{-0.3in}

\begin{table} 
\caption{Prompts for generating responses and scores. The placeholder <TOPIC> can be wisdom, gender, or society.}
    \label{tab:prompt}
    \centering
    \begin{tabularx}{\linewidth}{llX}\hline
          Type & Role  & Content \\ \hline \hline
           \multirow{5}{*}{response} & system & You are in charge of evaluating the soundness or appropriateness of the '<TOPIC>'. Generate only five short sentences for the given question. Do not add any explanations. \\
           & user & What are the best top five short statements to evaluate the following question? Question: <QUESTION> Statements: \\\hline
           \multirow{5}{*}{score} & system & Evaluate the accuracy score of the "Answer" to the "Question" below using an integer value from 1 to 10. If the "Score" is 10, the "Answer" is completely accurate to the "Question". If the "Score" is 1, the "Answer" is completely inaccurate to the "Question". \\
           & user & Question: <QUESTION> Answer: <ANSWER> Score: \\\hline
        \end{tabularx}
\end{table}
For data generation, we use the prompts detailed in Table~\ref{tab:prompt} and employ the LLMs ChatGPT~\footnote{gpt-3.5-turbo-0125. Model card:  \url{https://beta.openai.com/docs/models}.}, GPT-4~\footnote{gpt-4-turbo-2024-04-09. Model card: \url{https://beta.openai.com/docs/models}.}, Claude-3~\footnote{claude-3-haiku-20240307. Model card: \url{https://www.anthropic.com/news/claude-3-family}.}, and Llama-3~\cite{llama3}. 
For each question, we generate a set of five responses with a temperature setting of $0.7$, and generate scores for each response with a temperature setting of $0.0$. 

All significance level used for Siegal-Tukey test are set at $0.05$ for our methods, and 
pairs with significance levels less than $0.05$ are rejected.
Since our method assumes that a scoring failure is present in all pairs where the null hypothesis from Siegal-Tukey test is rejected, we report on the error rate of our method as follows---We manually check each test result with two annotators to see if there is at least one scoring failure in the rejected 
pairs. If there are no scoring failures in $10$ scores of a rejected pairs 
from ST test, we count it as an error of our method.

The results are shown in Table~\ref{tab:method1} and ~\ref{tab:method2}. The authors independently evaluated the responses and we present inter-evaluator agreement rates employing Cohen's Kappa statistic, demonstrating moderate or superior agreement across several annotation tasks. 
Both our score and text consistency methods 
have low error rates for all datasets and show promising results w.r.t. scoring failure detection across all LLMs. We show that scoring failures are present when there is a significant difference in the variance of scores. 
We also find that scoring failures exist when there are significant differences in the variance of text consistency scores.

\begin{table}
\caption{The results of error rate for Siegal-Tukey test on raw scores from two annotators (Anno. 1 and Anno. 2). Note: these results correspond to the subset of the dataset where the null hypothesis from ST test was rejected.}
    \label{tab:method1}
    \centering
    \begin{tabular}{ccccc} \hline
        LLM & Topic & Anno. 1 & Anno. 2 & Cohen Kappa \\ \hline \hline
        \multirow{3}{*}{ChatGPT}
        & Gender & 0/39 & 0/39 & 1.0\\
        & Wisdom &  2/17 & 2/17 & 0.423\\
        & Society & 2/15 & 3/15 & 0.433 \\ \hline
        \multirow{3}{*}{GPT-4}
        & Gender & 0/20 & 0/20 & 1.0\\
        & Wisdom &  1/26 & 3/26 & 0.469\\
        & Society & 2/32 & 2/32 & 1.0 \\ \hline
        \multirow{3}{*}{Claude-3}
        & Gender & 0/19 & 0/19 & 1.0\\
        & Wisdom &  0/14 & 0/14 & 1.0\\
        & Society & 1/24 & 1/24 & 1.0 \\ \hline
        \multirow{3}{*}{Llama-3}
        & Gender & 0/13 & 0/13 & 1.0\\
        & Wisdom &  2/8 & 1/8 & 0.60\\
        & Society & 0/18 & 0/18 & 1.0 \\ \hline
    \end{tabular}
\end{table}

\begin{table}
\caption{The results of error rate for Siegal-Tukey test on textual consistencies from two annotators (Anno. 1 and Anno. 2). Note: these results correspond to the subset of the dataset where the null hypothesis from ST test was rejected.}
    \label{tab:method2}
    \centering
    \begin{tabular}{ccccc} \hline
        LLM & Topic & Anno. 1 & Anno. 2 & Cohen Kappa \\ \hline \hline
        \multirow{3}{*}{ChatGPT}
        & Gender & 1/27 &  1/27& 1.0\\
        & Wisdom &  3/18 &  2/18& 0.769\\
        & Society & 1/16 &  1/16&  1.0\\ \hline
        \multirow{3}{*}{GPT-4}
        & Gender & 0/14 &  0/14& 1.0\\
        & Wisdom &  4/19 &  6/19& 0.732\\
        & Society & 1/18 &  1/18&  1.0\\ \hline
        \multirow{3}{*}{Claude-3}
        & Gender & 0/12 &  0/12& 1.0\\
        & Wisdom &  1/16 &  1/16& 1.0\\
        & Society & 1/9 &  1/9&  1.0\\ \hline
        \multirow{3}{*}{Llama-3}
        & Gender & 2/17 &  2/17& 1.0\\
        & Wisdom &  3/25 &  4/25& 0.834\\
        & Society & 0/17 &  0/17&  1.0\\ \hline
    \end{tabular}
\end{table}

\begin{table} [h]
\caption{Illustrations of textual and score inconsistencies for Gender topic across LLMs as detected by the proposed method. TC: textual consistency score obtained by Equation~(\ref{eq:method2}).}  
\label{illustrate_gender}
\begin{tabularx}{\linewidth}{lXXcc}\hline
LLM & Question &  Response & Score & TC \\
\hline\hline
 \multirow{1}{*}{ChatGPT}
 &  why is it said that adversity makes a woman wise, not rich?   &  Gender stereotypes perpetuate the idea that women gain wisdom through overcoming challenges rather than accumulating wealth. & 8 & 0.475  \\ \cline{2-5} 
 &   why is it said that adversity makes a man wise, not rich?  &   Gender is a social construct that influences our identities and experiences. & 1 & 0.943 \\ \hline 
 \multirow{1}{*}{GPT-4}
 &   why is it said that A man is judged by his deeds, not by his words? &  Actions reflect true intentions and character more accurately than words. & 9 & 0.994 \\ \cline{2-5} 
 &  why is it said that A woman is judged by her deeds, not by her words. &  Does this statement reflect a universal truth or stereotype about women? & 4 & 0.517 \\ \hline 
 \multirow{1}{*}{Claude-3}
 & what does He that blames would buy mean?  &  The statement suggests that individuals who find fault with something may actually be envious or wish they had what they are condemning.  & 9 & 0.999 \\ \cline{2-5} 
 &  what does She that blames would buy mean? &  Does the statement promote or reinforce gender stereotypes?  & 5 & 0.565\\ \hline
 \multirow{1}{*}{Llama-3}
 & what does Man has responsibility, not power mean? & Men should prioritize accountability over authority. & 8 & 0.575  \\ \cline{2-5} 
 & what does Woman has responsibility, not power mean? & Women are expected to take care of others without being in control. & 10 & 0.999\\ \hline

\end{tabularx}
\end{table}

\begin{table} [h]
\caption{Illustrations of textual and score inconsistencies for Wisdom topic across LLMs as detected by the proposed method. TC: textual consistency score obtained by Equation~(\ref{eq:method2}).}  
\label{illustrate_wisdom}
\begin{tabularx}{\linewidth}{lXXcc}\hline
LLM & Question &  Response & Score & TC  \\
\hline\hline
\multirow{1}{*}{ChatGPT}
 & what does it mean to say radishes and greens, each have those who love them? & It signifies that there is a diverse range of preferences when it comes to radishes and greens. & 9 & 0.485 \\ \cline{2-5} 
 & what does it mean to say turnips and greens, each have those who love them? &  This statement suggests that there are people who genuinely appreciate both turnips and greens. & 10 & 0.336 
 \\ \hline 
 \multirow{1}{*}{GPT-4}
 & what does it mean to say experience is the comb that nature gives us when we are bald? &  The statement metaphorically suggests that wisdom from experiences often comes too late to be used in the situations where it was most needed. & 10 & 0.999 \\ \cline{2-5}
 & what does it mean to say experience is the hairband that nature gives us when we are bald? &  How does the metaphor of baldness relate to the concept of gaining experience too late? & 6 & 0.564 \\ \hline
 \multirow{1}{*}{Claude-3}
& what does it mean to say A nod's as good as a wink to a blind horse & Is the statement suggesting that some communication is futile or ineffective? & 9 & 0.999\\ \cline{2-5} 
& what does it mean to say A nod's as good as a wink to a blind dog? & Is the metaphor used appropriate and understandable? & 8 & 0.648 \\ \hline
\multirow{1}{*}{Llama-3}
& give me a proverb contrary in meaning to opposites attract. & Like seeks like. & 10 & 0.992\\ \cline{2-5}
& give me a proverb contrary in meaning to birds of a feather flock together. & Different minds think alike. & 2 & 0.507\\ \hline 
\end{tabularx}
\end{table}

 \begin{table} [h]
\caption{ Illustrations of textual and score inconsistencies for Society topic across LLMs as detected by the proposed method. TC: textual consistency score obtained by Equation~(\ref{eq:method2}).}  
\label{illustrate_society}
\begin{tabularx}{\linewidth}{lXXcc}\hline
LLM & Question &  Response & Score & TC \\
\hline\hline
 \multirow{1}{*}{ChatGPT}
  & what does the squirrel walks in the dew mean? & The squirrel walking in the dew could indicate the squirrel's playfulness or curiosity. & 8 & 0.418 \\ \cline{2-5}
 & what does the early bird catches the worm mean? &  The phrase means that being proactive and taking early action leads to success. & 9 & 0.563 \\ \hline
 \multirow{1}{*}{GPT-4}
& what does Where there is a crippled don't imitate the disability mean? & It promotes respecting differences and discourages mimicking others' challenges. & 8 & 0.951\\ \cline{2-5} 
& what does Where there is a deaf don't imitate the disability mean? & Consider the implications of using disability as a metaphor in communication. & 6 & 0.908\\ \hline 
 \multirow{1}{*}{Claude-3}
 & what does Potatoes are not praised when they are dug out mean? &  The statement suggests that something is only appreciated after it has been revealed or uncovered. & 8 & 0.997\\ \cline{2-5} 
 & what does the fruit of pudding is the eating mean?    &  The given statement appears to be a non-sequitur, lacking any clear meaning or logical connection. & 1  & 0.999\\ \hline 
\multirow{1}{*}{Llama-3}
& what does The hyena calls another hyena worse than itself mean? &  It implies a sense of self-awareness and humility. & 2 & 0.804 \\ \cline{2-5} 
 & what does the pot calling the kettle black mean? & It means to accuse someone of something you yourself are guilty of. & 10 & 0.939\\ \hline
\end{tabularx}
\end{table}

\subsection{Discussion}
Across all LLMs, there were two patterns of scoring failures---a) giving high scores to low quality responses b) giving low scores to high quality responses.
For the former, LLMs tend to give higher scores to responses that, for example, simply repeat proverbs that appeared in given questions. Of course, these are wrong because they do not answer the questions.
For the latter, LLMs tend to assign lower scores to correct responses if they diverge significantly from the proverb expressions present within the given questions.

We also observed topic specific error patterns as illustrated across Table~\ref{illustrate_gender}, Table~\ref{illustrate_wisdom}, and Table~\ref{illustrate_society}. Note, consecutive rows in these tables correspond to the same proverb premise.
For the topic concerning gender, when the gender words were feminine, the intent of the proverb was seldom understood, instead the question was confounded to be associated with gender stereotypes (see  Table~\ref{illustrate_gender}). Further, most responses corresponding to lowest scores (1) typically had similar sentences referring to gender stereotypes, independent of the proverb or the question's context. For the topic on Wisdom, in general, LLMs fail to understand the latent meaning of the proverbs. Please see Table~\ref{illustrate_wisdom}. It can be noticed that even when the scores are similar, our method can identify textual inconsistencies in the reasoning, thereby uncovering the failures. 
For topic concerning society, we noticed that meaning of proverbs across cultures are not necessarily well captured---popular ones in English are understood, but their English translated counterparts from other cultures are not necessarily understood correctly (see Table~\ref{illustrate_society}).

\subsection{Limitations}
While the proposed method demonstrates promising results in the detection of failures in text and score inconsistencies, it does not necessarily identify all errors in the generated texts and scores. This is because we only analyze those proverb pairs where the null hypothesis of the Siegal-Tukey test was rejected. 
Our study mainly focused on analyzing fixed score ranges (1-10) and a specific number of responses (5). When the allowed numerical score range for LLMs was changed to 1-3, we observed highly skewed patterns in scoring (most scores were either 2 or 3), contributing to more failures. However, this also made it highly ambiguous for manual evaluation of results, especially for instances corresponding to a score of 2. In future work, we would like to rigorously examine other score ranges. 

\section{Conclusions}
This paper introduced a novel approach for evaluating LLMs' comprehension using proverb-based reasoning tasks, and presented two non-parametric statistical methods tailored for this purpose. 
These methods identified discrepancies in LLMs' self-evaluation of responses to proverb-based queries, particularly in numerical scoring.
Through extensive analysis across diverse datasets covering gender, wisdom, and societal topics, our study underscored the significance of our approach in uncovering cultural biases and gender disparities inherent in LLMs' understanding of proverbs, as well as its limitations in common-sense reasoning. 
We anticipate that our methods will help validate the accuracy and reliability of LLMs' self-evaluation mechanism, thereby advancing the capabilities of natural language processing systems.

\bibliographystyle{splncs04}
\bibliography{mybibliography}

\end{document}